# LiDAR-guided Stereo Matching with a Spatial Consistency Constraint

**Yongjun Zhang [a, *, 1], Siyuan Zou [a, 1], Xinyi Liu [a, *], Xu Huang [b], Yi Wan [a], and Yongxiang Yao [a]**

a     School of Remote Sensing and Information Engineering, Wuhan University, Wuhan 430079, China
b     School of Geospatial Engineering and Science, Sun Yat-Sen University, Zhuhai 519082, China
\*     Corresponding authors.
[1]     Both authors contributed equally to this manuscript.
E-Mail: zhangyj@whu.edu.cn (Y. Zhang), zousiyuan3s@whu.edu.cn (S. Zou), liuxy0319@whu.edu.cn (X. Liu), huangx358@mail.sysu.edu.cn (X. Huang), yi.wan@whu.edu.cn (Y. Wan), yaoyongxiang@whu.edu.cn (Y. Yao)

**Abstract:** The complementary fusion of LiDAR data and image data is a promising but challenging undertaking for generating high-precision and high-density point clouds. This paper proposes a LiDAR-guided stereo matching with a spatial consistency constraint framework, which considers the continuous disparity/depth changes in the homogeneous region of an image. First, the homogeneous pixels of each LiDAR projection point are defined based on their color/intensity similarity. Then, a riverbed enhancement function is proposed to optimize the cost volume of the LiDAR projection points and their homogeneous pixels for the purpose of matching robustness. Our formulation expands the constraint ranges of sparse LiDAR projection points that are adaptive to the percentage of LiDAR projection points on an image for optimizing the cost volume of pixels as much as possible. We applied our proposed method to Semi-Global Matching and AD-Census on both simulation and real datasets. When the LiDAR points percentage of the simulated datasets was 0.16%, the matching accuracy of our method achieved a sub-pixel level, while the original stereo matching algorithm was 3.4 pixels. Our experimental results show that our method provides good transferability across different scenarios and dense matching strategies. Furthermore, the qualitative and quantitative evaluations demonstrate that the proposed method is superior to two state-of-the-art cost volume optimization methods, especially in reducing mismatches in difficult matching areas and refining the boundaries of objects or buildings.

# 1. Introduction

Three-dimensional (3D) information perception is the digitization of real-world spatial information (Landeschi, 2018), which plays an important role in the areas of cultural heritage (Xiao et al., 2018), robotic navigations (Nefti-Meziani et al., 2015), autonomous driving (Arnold et al., 2019), change detection (Zhou et al., 2020), and building extraction (Liu et al., 2020), among others.

Automatic 3D reconstruction of objects and scenes at different scales is currently performed using range or image data. Active or range sensors, such as structured light scanners or laser scanners, are a common source of dense point clouds due to their ease of use, non-contact, and ability to capture millions of points in a very short period of time (Remondino et al., 2014). However, structured light scanners commonly fail in the presence of sunlight and provide a limited sensing range (Salvi et al., 2004), and laser scanners can obtain only low-density point clouds when funding is limited. Compared with Light Detection and Ranging (LiDAR), the image-based approach is a passive, pixel-wise, low-cost technology for obtaining the 3D structure of a scene that has a higher density and a finer topography. In difficult matching areas, such as shadows, low textures, and repeated textures, the image-based approach is often inaccurate, while LiDAR point clouds respond better in the above areas (Maltezos et al., 2016). Therefore, the complementary data fusion of LiDAR data and image data is a promising solution for generating dense, accurate, texture-rich point clouds (Huang et al., 2018).

The fusion methods of LiDAR data and image data can have different levels depending on the differences in the emphasis of the data sources, which include: 1) point clouds fusion, 2) LiDAR data interpolation aided by images, and 3) dense image matching constrained by LiDAR data. Point clouds fusion directly integrates the LiDAR data and the point clouds generated by dense image matching, which plays an important role in geoscience applications, such as Digital Surface Model (DSM) generation (Schenk and Csathó, 2002), urban 3D modeling (Gao et al., 2020; Mandlburger et al., 2017), and forest resource management (White et al., 2013). However, the completeness and accuracy of the matching point clouds and LiDAR data are not significantly improved individually. LiDAR data interpolation aided by images can produce complete and finer point clouds with the guidance of image information (He et al., 2012), but it can also produce unreliable predictions in image areas that are lacking in point clouds. The more reliable way is dense image matching constrained by LiDAR data, which integrates the high fidelity of LiDAR data into the advanced dense image matching framework. The high fidelity of LiDAR data can be reflected in many aspects, such as reducing the disparity search range (Zhou et al., 2018), optimizing the matching cost (Poggi et al., 2019), and adjusting or adding the penalty parameters (Huber and Kanade, 2011). Applying the above strategies to LiDAR projection points can achieve better dense matching results when the point cloud is dense, but when the point cloud is too sparse, the improvement effect is weak. Expanding the number of LiDAR projection points by predicting the disparities of the neighboring pixels around the LiDAR projection points can overcome the above problems (Shivakumar et al., 2019). But this approach also brings a new challenge that the dense matching results rely on the accuracy of the predicted disparities.

We propose a method of LiDAR-guided stereo matching with a spatial consistency constraint that effectively can expand the constraint ranges of sparse LiDAR projection points without predicting the disparities of the homogeneous pixels. Our riverbed enhancement function (Subsection 3.2) in our proposed method makes full use of spatial consistency. Its basic assumption is that the continuous disparity/depth changes in the homogeneous region of an image. The main contributions of this paper are as follows:

(1) A riverbed enhancement function is proposed to expand the constraint ranges of sparse LiDAR projection points without predicting the disparities of the homogeneous pixels.

(2) Our proposed method is fully automatic and adaptive to the percentage of LiDAR projection points on an image. Even if the LiDAR projection points are very sparse, the adaptive parameters ensure that the matching accuracy of our method is greatly improved than the original stereo matching algorithm.

(3) Our proposed method is suitable for indoor, street, aerial, and satellite image datasets and provides good transferability across Semi-Global Matching (SGM) and AD-Census.

The remainder of this paper is organized as follows. Section 2 provides an overview of past related works. Section 3 describes our LiDAR-guided stereo matching process with a spatial consistency constraint in detail. Section 4 presents our experimental results on simulated and real datasets. Section 5 concludes this paper by highlighting our contribution to LiDAR-guided stereo matching and its future use.

## 2. Related Work

In the last several years, there have been many attempts to fuse LiDAR data and images for producing dense and accurate point clouds. One simple method is the direct integration of the LiDAR data and the point clouds generated by dense image matching. Since LiDAR data offer high precision but sparse features, many scholars have proposed obtaining dense reconstruction by sparse LiDAR data interpolation aided by images. Thanks to recent advancements in dense image matching, LiDAR constrained dense matching is receiving a great deal of attention as well. Therefore, based on the differences in the emphasis of the data sources in the fusion process, the fusion methods of laser range data and images can be categorized as follows: 1) point clouds fusion, 2) LiDAR data interpolation aided by images, and 3) dense image matching constrained by LiDAR data.

### 2.1 Point clouds fusion

Point clouds fusion directly integrates the LiDAR data and point clouds generated by dense image matching. The synergism of two different data types considerably exceeds the information obtained by individual sensors, which plays an essential role in many geoscience applications. This approach was typically used to detect changes and create more robust and complete digital terrain models by combining airborne LiDAR scanning data and aerial imagery (Schenk and Csathó, 2002). Combining with the higher density of the matching point clouds and the higher reliability of LiDAR point clouds produced an improved digital surface model, especially in narrow alleys and courtyards (Mandlburger et al., 2017). In forest management, integrating LiDAR data that can determine tree height is being used to create digital terrain models, and image data that can extract tree species and tree health information were being used to create a complete forest resource inventory model (White et al., 2013). In the field of heritage protection, combining LiDAR data and matching point clouds also was being used to create a high-resolution 3D model of monuments, statues, or facades (Schenk and Csathó, 2002). In the large-scale and complicated architectural scene, the images and laser scans were merged with a coarse-to-fine strategy (Gao et al., 2020) for reconstructing an accurate and complete 3D model (point cloud or surface mesh). Although point clouds fusion has broad application scenarios, the completeness and accuracy of the matching point clouds and the LiDAR data have yet to be significantly improved individually.

## 2.2 LiDAR data interpolation aided by images

LiDAR data interpolation aided by images combines LiDAR data with a single registered image to estimate the depths of all the points/pixels in higher-resolution grids after interpolation (Huang et al., 2018). The image-guided method assumes that most discontinuities in the image of one sensor correspond to discontinuities in the other sensors (Zomet and Peleg, 2002). The most usual image-guided method is the bilateral filter (Kopf et al., 2007)and its optimization method (Min et al., 2012). Weighted average based on homogeneous pixels is an effective and fast method (Wang and Ferrie, 2015). Several super-resolution reconstruction methods based on deep learning outperformed the interpolation method in accuracy and efficiency (Cui et al., 2021; Eldesokey et al., 2019; Zhang et al., 2019). However, LiDAR data interpolation aided by images is an unreliable predictor in an image area lacking in point clouds.

## 2.3 Dense image matching constrained by LiDAR data

Dense image matching constrained by LiDAR data first obtains the position of the laser point on the image, which can be regarded as a high-precision control point. Then, The constraints of the LiDAR data are integrated into the dense matching framework. Dense image matching constrained by LiDAR data is advantageous for automated 3D reconstruction due to the complementary characteristics of LiDAR and images (Nickels et al., 2003). Several studies used LiDAR data projected on an image as a seed point and then generated a depth/disparity map of the entire image through a region growing algorithm (Gandhi et al., 2012; Veitch-Michaelis et al., 2015). Other studies adjusted or added the penalty parameters in the cost aggregation process that depends on the distance between the current disparity and the disparity of the LiDAR projection point (Huang et al., 2018; Huber and Kanade, 2011). The above approach can achieve good results in specific dense image matching methods, but it is not very general for different dense matching algorithms. Reducing the disparity search range for each pixel in an image based on LiDAR data is a general strategy that can effectively improve the matching accuracy and efficiency (Huang et al., 2018; Zhou et al., 2018). The cost volume stores the matching cost of each pixel on the reference image with the matching image in the disparity range. Directly optimizing the cost volume can be applied to most of the stereo matching algorithms with cost volume. The Gaussian enhancement function was used to optimize the cost volume on each LiDAR projection point, which improved the accuracy of dense matching (Poggi et al., 2019). Furthermore, expanding the LiDAR projection points yielded much improved dense matching results by predicting the disparities of the neighboring pixels around the LiDAR projection points (Shivakumar et al., 2019); however, its matching accuracy relies on the accuracy of the predicted disparities.

To solve the above issues, we propose a novel method of LiDAR-guided stereo matching with a spatial consistency constraint that includes the following: 1) the spatial neighborhood homogeneous pixels of each LiDAR projection point are calculated and 2) the cost volume of these pixels is updated with a specific modulation function. This method can effectively extend LiDAR projection points without predicting the disparities of the homogeneous pixels and thereby generate a more robust and accurate disparity map.

## 3. Methodology

Dense image matching usually is conducted in the following four steps: cost computation, cost aggregation, disparity optimization/computation, and disparity refinement (Scharstein and Szeliski, 2002). Following the traditional dense image matching process, LiDAR-guided stereo matching with a spatial consistency constraint optimizes the cost volume calculated by cost computation by using external LiDAR projection points. The cost volume of LiDAR projection points and homogenous pixels are optimized without changing the traditional dense image matching pipeline. This method effectively extends LiDAR projection points without predicting the disparities of the homogeneous pixels, which belongs to dense image matching constrained by LiDAR data. The above method is abbreviated as our proposed riverbed method in this paper. Figure 1 shows the workflow of our riverbed method, where the black part is the traditional dense image matching pipeline, and the red part is the research focus of this paper. First, the homogeneous pixels around the LiDAR projection points are calculated from the image based on color/intensity similarity, which is described in detail in Subsection 3.1. Then, the cost volume is optimized through our novel riverbed enhancement function, which is defined in detail in Subsection 3.2.

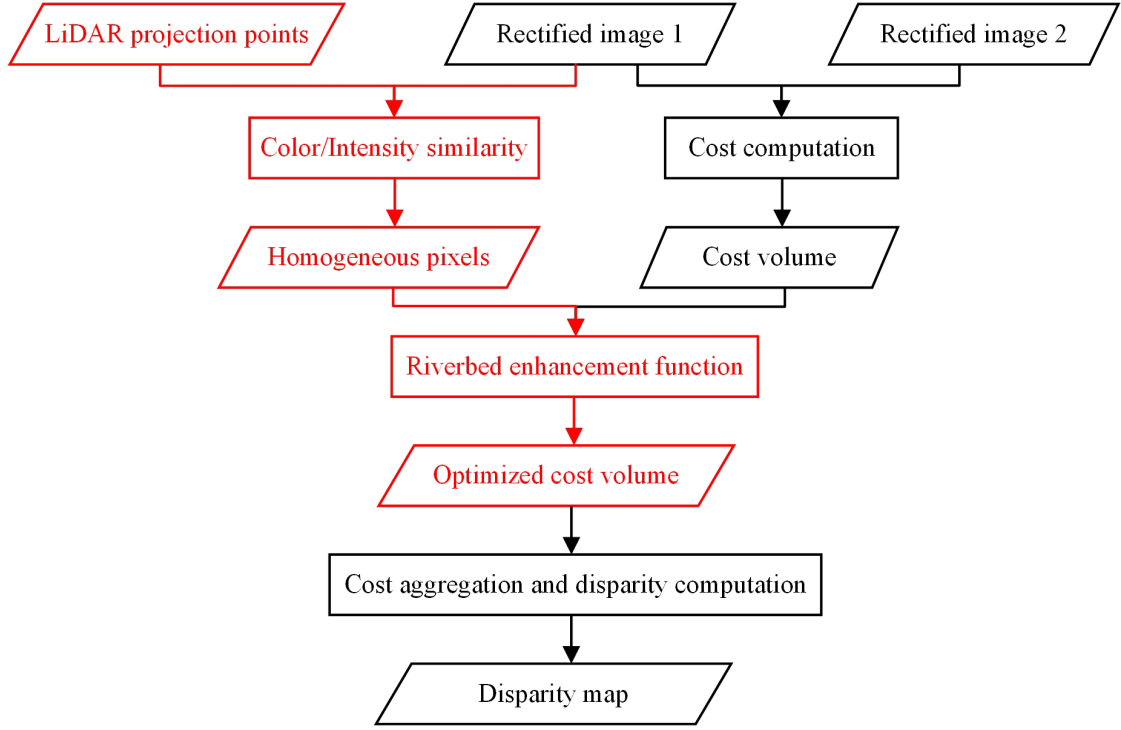

**Figure 1.** The workflow of our riverbed method

## 3.1 Homogeneous pixels around the LiDAR projection point

Our riverbed method defaults that all LiDAR projection points are correct. Registering the images and the laser range data is a prerequisite, and generating the LiDAR projection points on the image is the initial step of our riverbed method. For the purpose of brevity in this paper, all the unregistered data was accurately registered by using multi-features matching as a preprocess (Zhang et al., 2015). First, the outliers in the LiDAR data are filtered out. Then, the LiDAR data are projected to the left and right rectified images according to the precise orientation parameters of the image. In the process of projection, a simple and fast hidden point removal operator (Katz et al., 2007) is used to remove occluded point clouds. The coordinates of a LiDAR point on the left and right rectified images at this point are $(x_L, y_L)$ and $(x_R, y_R)$, respectively. Thus the true disparity of $(x_L, y_L)$ is $d_m$:

$$d_m = x_R - x_L \tag{1}$$

where $(x_L, y_L)$ is the LiDAR projection point, $d_m$ is the disparity value corresponding to the LiDAR projection point. The disparity value generated by dense matching should also be the same as $d_m$, which is the basic assumption of our riverbed method.

Homogeneous pixels are pixels with no significant change in intensity or color values within the spatial neighborhood of the LiDAR projection point. Inspired by the idea of a bilateral filter (Tomasi et al., 1998), homogeneous pixels are defined as pixels of color/intensity similarity within a certain distance to preserve the shape boundary of the image. We follow this hypothesis by creating a weighted Gaussian distribution based on the spatial and intensity proximities to the centric pixel:

$$W_{(i,j)} = 1 - \exp\left(-\frac{\|(i,j)-(x_L,y_L)\|^2}{2\sigma_{xy}^2} - \frac{(I(i,j)-I(x_L,y_L))^2}{2\sigma_I^2}\right) \quad (2)$$

where $(i, j)$ and $(x_L, y_L)$ are the positions of the current pixel and center pixel; $I(i, j)$ and $I(x_L, y_L)$ are the intensity of the current and the centric pixel, respectively. The exponent part is always negative, and the value range of $W_{(i,j)}$ is [0-1]. Given the threshold $\gamma$, we can obtain similar pixels around the LiDAR projection point:

$$S(x_L, y_L) = \{(i,j) \in Window(x_L, y_L) \mid W_{(i,j)} > \gamma\} \quad (3)$$

where $S(x_L, y_L)$ stores the homogeneous pixels around the LiDAR projection point. $Window(x_L, y_L)$ refers to a square with a center pixel of $(x_L, y_L)$ and a window size of $s \times s$. The illustration of homogeneous pixels is shown in Figure 2.

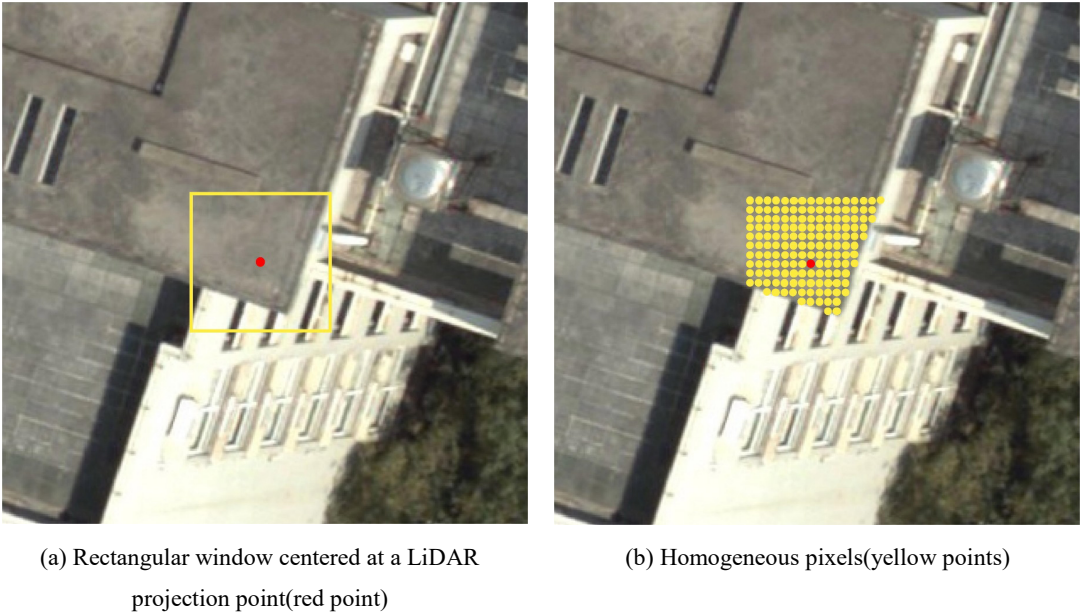

(a) Rectangular window centered at a LiDAR projection point(red point)

(b) Homogeneous pixels(yellow points)

**Figure 2.** Illustration of homogeneous pixels

Intuitively, the larger the window size is, the more homogeneous pixels there will be near the $\sigma_{xy}$ and $\gamma$ can ensure that when the pixels are far away from the LiDAR projection point, the pixels with the same intensity/color will not be judged as homogeneous pixels. Section 4 will discuss the recommended values of all the parameters, including the window size in the formula. When the density of LiDAR points is too high, some pixels on the image may be homogeneous pixels around multiple LiDAR projection points. To ensure that all the pixels in the image are the homogenous pixels of at most one LiDAR projection point, we follow two basic rules: 1) $Window(x_L, y_L)$ cannot contain other LiDAR projection points and 2) a matrix the same size as the image is created to ensure that each pixel is at most a homogeneous pixel of a LiDAR projection point.

### 3.2 LiDAR constraint cost volume of stereo matching

The first step of cost computation is to obtain the feature description of each pixel on the left and right images. Then, the matching cost is calculated by measuring the distances of the feature description of the matching point:

$$C((x,y),d) = Dist(I(x,y), J(x+d,y)) \qquad (4)$$

where $C((x,y),d)$ represents the matching cost of the current pixel $(x,y)$ with $d$ as the disparity. $I(x,y)$ and $J(x+d,y)$ are the feature descriptions of the matching points of the reference image and the matching image, respectively. $Dist$ is the similarity measure of the feature descriptions of the matching points, such as the Hamming distance. The cost volume stores the matching cost of each pixel on the reference image with the matching image in the disparity range, which has dimension $H \times W \times D$, with $H \times W$ being the resolution of the reference image and $D$ being the maximum disparity search range. The cost volume runs through cost computation, cost aggregation, and disparity optimization/computation, which has a direct impact on the results of dense matching. In this paper, we abbreviate the 3D cost volume of the entire image as $F$.

Optimizing the cost volume of LiDAR projection points based on the high fidelity of LiDAR data is an effective way to improve matching accuracy. Gaussian enhancement function is a very good method, which fully considers that the closer to the real disparity of the LiDAR projection point, the higher the confidence of the disparity $d$ Poggi et al., 2019). Gaussian enhancement function is used to optimize the cost of all disparities within the disparity search range of the LiDAR $d_m$ calculated by the LiDAR data and its neighborhood disparities should be reduced, and the cost of other disparities within the disparity search range should be increased, as shown in Figure 3 (a). Gaussian enhancement function optimizes the cost volume by selecting a modulation function with a height of $k$ and a standard error of $c$, and thereby produces an enhanced cost volume $G$ from the initial cost volume $F$:

$$G = [k*(1-e^{-\frac{(d-dm)^2}{2c^2}})]*F \qquad (5)$$

Although the Gaussian enhancement function improves the result of dense matching, when the LiDAR points are too sparse, the improved effect will be quite limited. Therefore, just optimizing the cost volume of LiDAR projection points is far from enough.

A general hypothesis is that the continuous disparity/depth changes in the homogeneous region of an image, which has been successfully utilized by many powerful matching algorithms (Hirschmuller, 2005; Huang et al., 2020; Qin, 2017; Yoon and Kweon, 2006). In the process of LiDAR constrained dense matching, LiDAR data not only can be used to constrain the cost volume of the pixel where the LiDAR projection point is located but also the cost volume of the homogeneous neighborhood pixel of the LiDAR projection point. Although several studies also consider the neighborhood points near the LiDAR projection point pixel, its modulation function depends on too many parameters and has nothing to do with the initial cost volume (Shivakumar et al., 2019). To make full use of the LiDAR constraints, the neighborhood information of the LiDAR projection points, and the initial cost volume, this paper introduces a more reliable updating method of cost volume.

As shown in Figure 3(b), a function that resembles a riverbed is proposed, which is called the riverbed enhancement function, taking into account the spatial consistency of homogeneous pixels. First, we calculate the similarity $W_{(i,j)}$ $w$ between the neighborhood pixel and the LiDAR projection point. When the disparity of the LiDAR projection point is $d_m$, the disparity of each homogeneous pixel has higher confidence between $d_m - w$ and $d_m + w$, while lower confidence in other disparities. Thus, the cost volume is uniformly adjusted within the disparity range $(d_m - w, d_m + w)$, while the cost volume is continuously enhanced with Gaussian outside the above range.

$$G = \begin{cases} [k*(1-e^{-\frac{(d-d_m+w)^2}{2c^2}}) + W(i,j)]*F & d \leq d_m - w \\ W(i,j)*F & d_m - w < d < d_m + w \\ [k*(1-e^{-\frac{(d-d_m-w)^2}{2c^2}}) + W(i,j)]*F & d \geq d_m - w \end{cases} \quad (6)$$

In the above formula, the similarity of the LiDAR projection points is added to each piecewise function, which effectively guarantees the continuity of the piecewise function. The riverbed enhancement function is equivalent to the Gaussian enhancement function when the pixel is located at the LiDAR projection point (the Euclidean distance $w = 0$ and the similarity $W_{(i,j)} = 0$). Therefore, the Gaussian enhancement function is a special case of the riverbed enhancement function. This model is not only an extension of the Gaussian model, but it also makes full use of the continuous disparity/depth changes in the homogeneous region of an image.

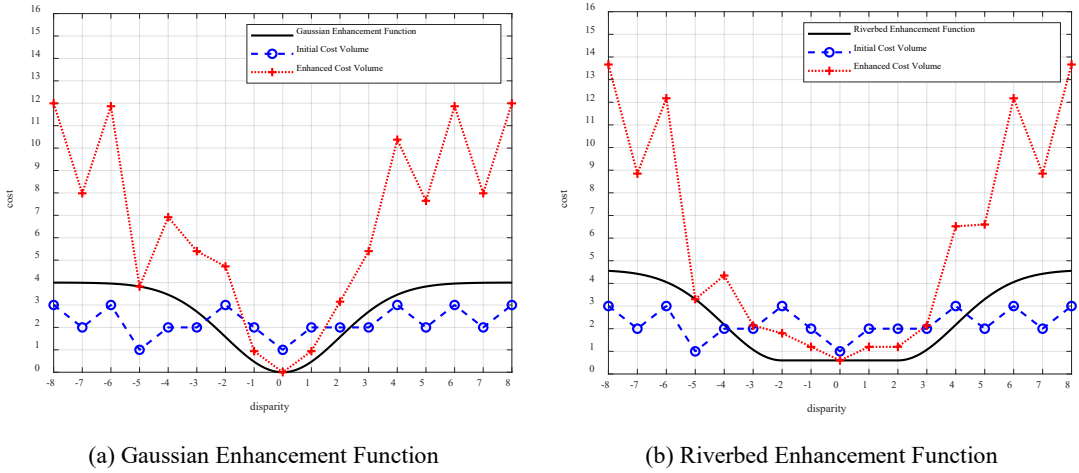

(a) Gaussian Enhancement Function    (b) Riverbed Enhancement Function

**Figure 3.** Modulation function used for cost volume updating: (a) Gaussian enhancement function is used to enhance the cost volume of LiDAR projection points and (b) our riverbed enhancement function is used to enhance the cost volume of LiDAR projection points and their homogeneous pixels.

## 4. Experiments

### 4.1. Experimental settings

The inputs of our riverbed method are rectified images and LiDAR data. The performance of our riverbed method was discussed on various stereo datasets, including indoor, street, satellite, and aerial images. The overview of the input data is shown in Table 1. In addition to stereo images, the street and aerial stereo image datasets provide LiDAR data, but the indoor and satellite stereo image datasets provide reference disparity maps collected by structured lighting scanners and fused DSM, respectively. According to whether the LiDAR data source contains real LiDAR data, the above datasets are divided into real datasets and simulated datasets. On simulated datasets, we simulated LiDAR projection points by randomly sampling the reference disparity map. Different sampling intervals mean different LiDAR points percentages, which helps to verify the degree of influence of our riverbed method on different LiDAR points percentages. LiDAR points percentage refers to the number of pixels of the LiDAR projection points divided by the number of pixels in the image area where the LiDAR projection points are located.

**Table 1.** Datasets overview

| Dataset | Scenes | LiDAR data source | Classification | LiDAR points percentage |
|---|---|---|---|---|
| Middlebury 2014 (Scharstein et al., 2014) | Indoor | Structured lighting scanners | Simulated Dataset | 0.05%, 0.08%, 0.16%, 0.44%, 1.23%, 11.11% |
| SatStereo (Patil et al., 2019) | Satellite | Fused DSM | Simulated Dataset | 5% |
| KITTI 2015 (Menze et al., 2015) | Street | LiDAR | Real Dataset | 5% |
| Guangzhou Stereo | Aerial | LiDAR | Real Dataset | 3% |

For our experiments, we integrated our riverbed method into Semi-Global Matching (SGM) (Hirschmuller, 2008) in Subsections 4.2 and 4.3, considering SGM offers a very good trade-off between matching accuracy and computational efficiency. Based on SGM (Li, 2020), the cost calculation method was census (Zabih and Woodfill, 1994), and the disparity refinement only used median filtering and did not use a left/right consistency check. Integration, in this case, refers to the fact that our riverbed method was used to optimize the cost volume calculated by the cost calculation. We compared our method to two other state-of-the-art cost volume optimization methods: the Gauss method (Poggi et al., 2019) and the Diffuse Based method (Shivakumar et al., 2019). The Gauss method only optimizes the cost volume of LiDAR projection points, and the Diffuse Based method optimizes the cost volume of the LiDAR projection points and neighboring pixels based on the predicted disparity map. Furthermore, we replaced SGM with AD-Census (Mei et al., 2011) in Subsection 4.4 to prove the migration of our riverbed method.

In the experimental configuration, all the experimental parameters were fixed, except that the window size changed according to the LiDAR points percentage. Similar to the fast bilateral solver, the default value of $\sigma_{xy}$ and $\sigma_l$ in bilateral filtering was 8 (Barron and Poole, 2016). Corresponding to the above parameters, the empirical value of the threshold $\gamma$ was 0.3. Existing experimental results show that the recommended values of $k$ and $c$ in the modulation function were 10 and 1 (Poggi et al., 2019).

The accuracy of the dense matching results was calculated according to the remaining LiDAR projection points (the unsampled pixels in the reference disparity map). The average error is the average value of the dense matching disparity $d_i$ and the true disparity $d_m$ at the position of the remaining LiDAR projection points. The definition of average error is as follows:

$$Avg = \frac{\sum_{i}^{N}|d_i - d_m|}{N} \qquad (7)$$

where $N$ is the number of remaining LiDAR projection points. The amount of outliers is also a commonly used accuracy indicator, which is defined as the percentage of the number of absolute differences greater than the 1/2/3 pixel between $d_i$ and $d_m$ divided by the number of remaining LiDAR projection points $N$.

### 4.2. Experimental results of simulated datasets

We evaluated our riverbed method first on the Middlebury 2014 dataset (Scharstein et al., 2014), which provides 23 stereo pairs of indoor scenes and corresponding ground truth data. The 5% of the ground truth data we randomly sampled as the LiDAR constraints and the remaining 95% of the ground truth data were used to evaluate the matching accuracy. The sampling percentage is consistent with the experiment of the Gauss method, which contributes to the fairness of comparison. Because the disparity refinement did not utilize a left/right consistency check, only the LiDAR projection points distributed on the left rectified image were used. To ensure that the cost volume of each pixel was optimized as much as possible, the window size of our riverbed method was set at 5×5.

A few example disparity maps are shown in Figure 4. In this figure, (a) is the left rectified image of stereo matching, (b) is the simulated LiDAR projection points used to constrain stereo matching, (c) is the matching result of the standard SGM, (d-f) are the matching results of the SGM with different LiDAR constraint strategies, and (g) is the ground truth disparity map. The standard SGM substantially restores the 3D shape of the object, as shown in Figure 4(c). However, the disparity map generated by SGM is worse at the edge of the object, mainly because the penalty parameters of the cost aggregation were fixed. With the introduction of LiDAR constraints, the edge of the disparity map generated by dense matching improved significantly, as shown in Figure 4(d-f). This phenomenon was more pronounced in our riverbed method and the Diffuse Based method because both methods extend the LiDAR constraints to homogeneous pixels. As shown in the white ellipse area in Figure 4, the disparity map generated by our riverbed method is more consistent with the ground truth disparity map and produces the least mismatches, thus reaffirms our observation that our riverbed method improves the matching robustness effectively.

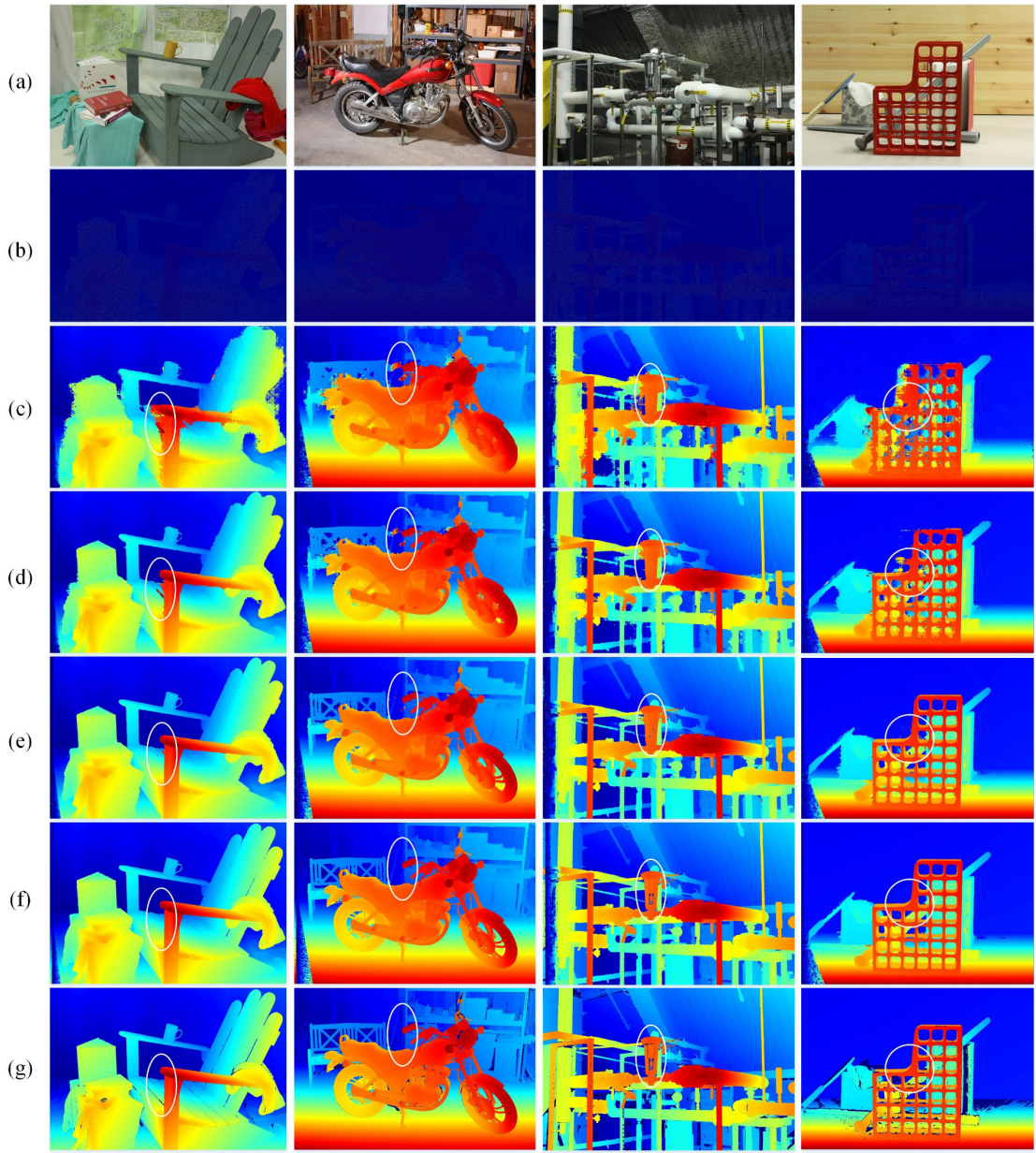

**Figure 4.** Experimental results on Middlebury 2014 dataset. (a) Left rectified image; (b) 5% random ground truth disparities; (c) SGM results (Hirschmuller, 2008); (d) Gauss results (Poggi et al., 2019); (e) Diffusion Based results (Shivakumar et al., 2019); (f) Our riverbed results; (g) Ground truth disparities.

In accordance with Section 3, the window size is the only parameter specified in our riverbed method. The window size is the edge length of a square centered on each LiDAR projection point, which limits the initial range of the homogeneous pixels. The window size is usually odd numbers, such as $3 \times 3$, $5 \times 5$, $7 \times 7$, $9 \times 9$, $11 \times 11$, $13 \times 13$, $15 \times 15$, $17 \times 17$, and $19 \times 19$. The window size cannot be set too large, mainly because the continuous disparity changes of homogeneous pixels are effective in a limited range. To take full advantage of the constraints of LiDAR data, the optimal solution is that the cost volume of each pixel can be optimized. To achieve this goal, the window size multiplied by the LiDAR points percentage should be greater than or equal to 1 as much as possible. Therefore, we randomly sampled 1:3 × 3, 1:9 × 9, 1:15 × 15, 1:25 × 25, 1:35 × 35, 1:45 × 45, and 1:∞×∞ of the ground truth data as the LiDAR constraints. Among them, 1:3 × 3 indicated that the LiDAR points percentage was 11.11%, and 1:∞×∞ indicated that there were no LiDAR points.

Figure 5 shows the relationship between window size and average error in a variety of LiDAR points percentages. As shown by the lowest line where the LiDAR points percentage was 1:3 × 3(11.11%) in Figure 5, we marked the average error corresponding to different window sizes. When the window size was 5×5, the average error was the smallest. As the window size increased, the average error gradually increased and tended to stabilize. When the LiDAR points percentage was 0.16%, the matching accuracy of our method achieved a sub-pixel level, while the standard SGM was 3.4 pixels. Even though the LiDAR points percentage was 1:45 × 45 (0.05%), the average error of our riverbed method was still more than 50% lower than the standard SGM. As shown by the lowest average error of all the LiDAR points percentage lines, we found the following: 1) the accuracy of the dense matching results increased with the increases in the LiDAR points percentage and 2) when the window size was slightly larger than the ratio of the image to the LiDAR resolution, the average error tended to stabilize and remained at a low level. Theoretically, the window size multiplied by the LiDAR points percentage was approximately equal to 1 to realize an optimized volume of each pixel on the image. It can be seen from Figure 5 that the window size multiplied by the LiDAR points percentage needs to be slightly greater than 1, which takes into account the uneven distribution of the LiDAR points and homogeneous pixels. This not only ensures that the cost volume of each pixel is optimized as much as possible but also ensures that the LiDAR projection point only optimizes the cost volume of the homogeneous pixels.

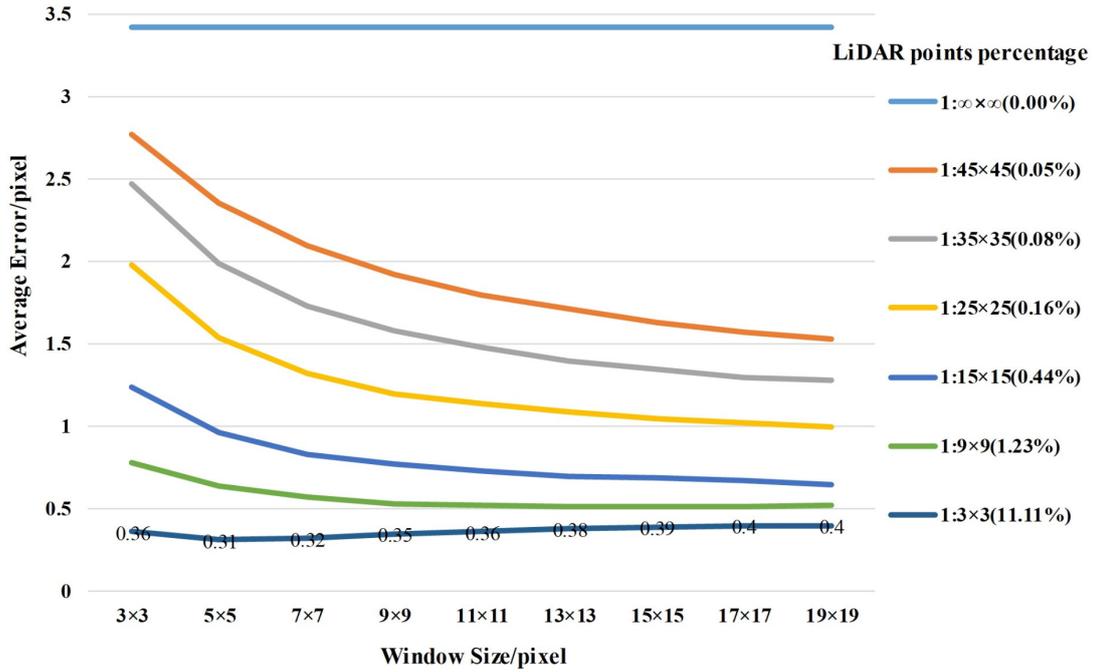

**Figure 5.** The relationship between window size and average error in a variety of LiDAR points percentages

To verify the effectiveness of our riverbed method on satellite stereo pairs, we used the satellite stereo pair dataset provided by Purdue University (Patil et al., 2019). This dataset consists of a set of stereo rectified image pairs photoed by WorldView-2 and WorldView-3 and presents the ground truth disparities for each pixel in a reference image. The disparities are ground truth by first constructing a fused DSM from the stereo pairs and then aligning the LiDAR with the fused DSM. However, the ground truth disparity map still retained a small part of the outlier, which was more common at the edges of buildings.

We randomly sample 5% of the ground truth disparities as LiDAR constraints, and the remaining 95% of the ground truth data were used to evaluate the matching accuracy, which is shown in Table 2. A few example disparity maps are shown in Figure 6. In standard SGM, there is no obvious disparity change at the edge of the buildings. There were slightly improved building edges in the Guided Stereo Matching results, but the results for our method and the Diffuse Based Method were closer to the ground truth. As can be seen in Table 2, the LiDAR constraint cost volume was significantly better than that of standard SGM in all metrics on the WorldView-2 and WorldView-3 datasets, and our riverbed method produced the best results. Our method reduced the amount of outliers by more than 50% compared to the original SGM and reduced the average error by 67%.

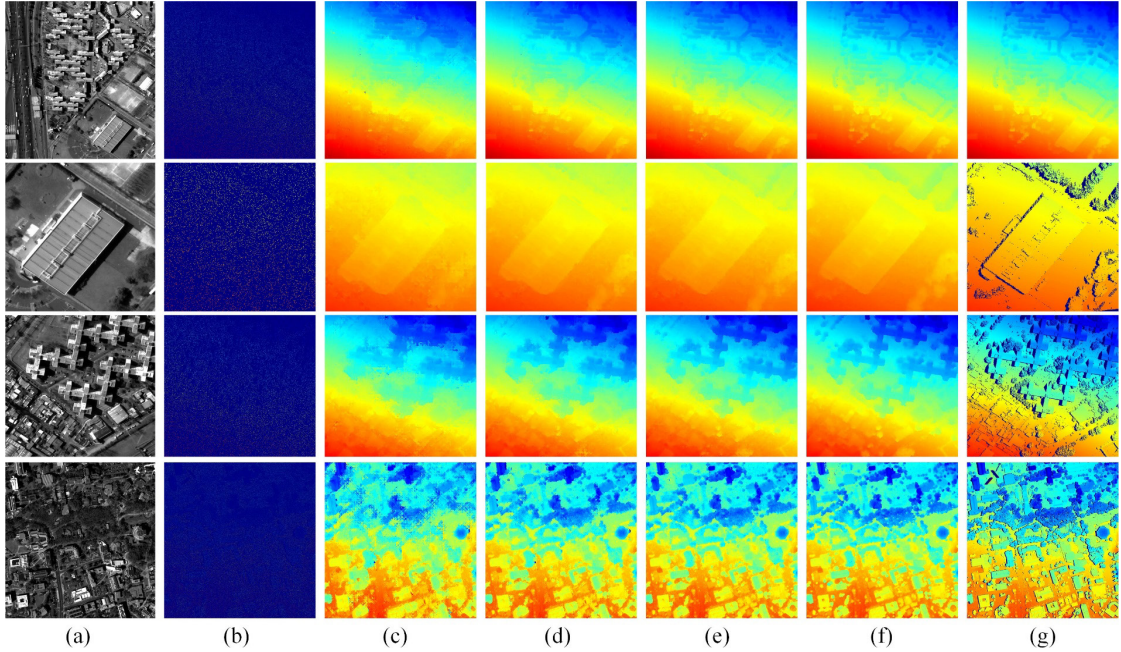

**Figure 6.** Experimental results on optical stereo satellite dataset. (a) Left rectified image; (b) 5% random ground truth disparities; (c) SGM results; (d) Gauss results; (e) Diffusion Based results; (f) Our riverbed results; (g) Ground truth disparities.

**Table 2.** The matching accuracy of standard and different LiDAR constraint SGM on WorldView-2 and WorldView-3 datasets.

| Method | WorldView-2 | | | | WorldView-3 | | | |
|---|---|---|---|---|---|---|---|---|
| | >1px | >2px | >3px | Avg/px | >1px | >2px | >3px | Avg/px |
| SGM | 70.41% | 40.40% | 29.02% | 3.01 | 55.80% | 36.45% | 28.39% | 4.47 |
| Gauss | 46.70% | 21.37% | 13.67% | 1.74 | 38.05% | 21.33% | 15.86% | 2.66 |
| Diffusion Based | 53.22% | 17.40% | 8.35% | 1.46 | 41.90% | 17.65% | 10.85% | 2.27 |
| **Riverbed** | **31.60%** | **8.76%** | **3.91%** | **0.98** | **26.17%** | **10.25%** | **5.88%** | **1.52** |

## 4.3. Experimental results of real datasets

We also evaluated the performance of our riverbed method on the KITTI 2015 dataset (Menze et al., 2015; Menze et al., 2018), which is comprised of 200 stereo pairs of street scenes and the corresponding ground truth data. However, the ground truth data are not pixel-wise in the grayscale image, but rather an accumulation of LiDAR points over a range of frames before and after the reference image. In our experiment, we used a subset of the ground truth data as our sparse depth input along with the stereo pair, and our evaluation on the ground truth data was conducted outside the sample set.

We randomly sampled 5% of the ground truth data as the LiDAR constraints, and the remaining 95% was used to evaluate the matching accuracy. According to the principle of parameter adaptation in Subsection 4.2, the window size was set at 5×5. A few example disparity maps are shown in Figure 7, which very clearly show that our riverbed method outperformed the other two methods. In particular, the disparity map generated by our riverbed method is smoother with fewer false matching points and has more accurate edges of objects, such as cars and tree trunks. Interestingly, the disparity map generated by our riverbed method was similar to that of the Diffused Based method. The disparity maps are similar because both methods expand the constraint range of the LiDAR projection points, but their results were different because they use different constraint strategies. In low light conditions (the first column of Figure 7), the results for the standard SGM algorithm were almost completely incorrect. However, all the LiDAR constrained cost volume methods generated more realistic disparity maps in the LiDAR area, and our riverbed method's results were the closest to the ground truth.

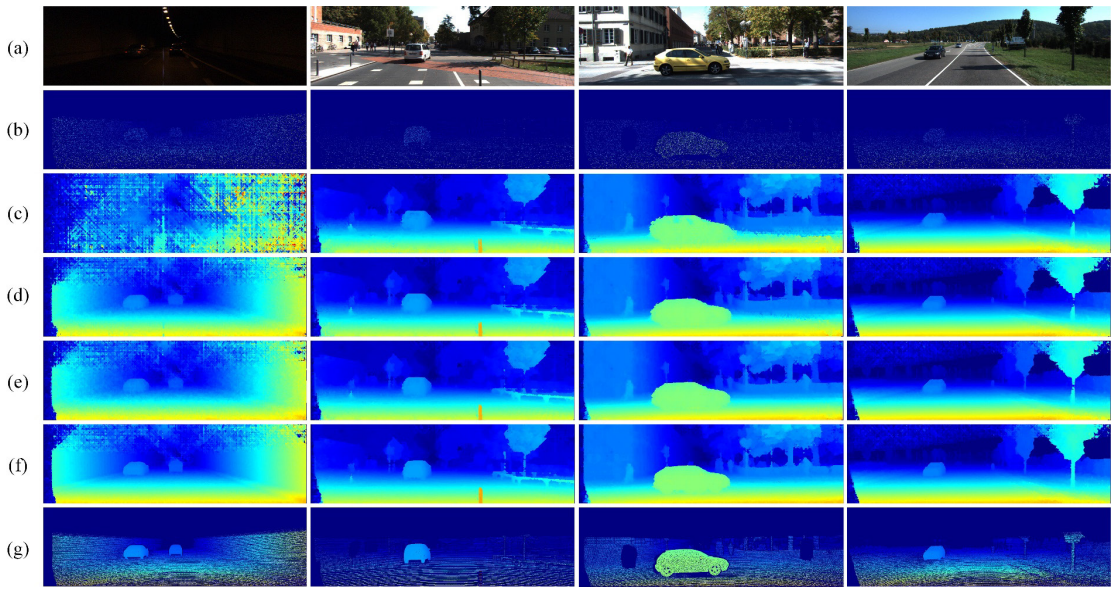

**Figure 7.** Experimental results on KITTI 2015 dataset. (a) Left rectified image; (b) 5% random ground truth disparities; (c) SGM results; (d) Gauss results; (e) Diffusion Based results; (f) Our riverbed results; (g) Ground truth disparities.

Table 3 shows the quantitative evaluation results for the KITTI datasets. Table 3 confirms that Gauss, the Diffusion Based method, and our riverbed method were significantly better than standard SGM in all the metrics, and our riverbed method produced the best results. On the KITTI 2015 dataset, our riverbed method also dramatically improved the amount of outliers and average error. In particular, compared to standard SGM, our riverbed method reduced the amount of outliers >1, >2, and >3 by more than 80%; and compared to Gauss and the Diffusion Based Method, it reduced the outliers by more than 50%.

Table 3. The matching accuracy of standard and different LiDAR constraint SGM on the KITTI datasets.

| Method | >1px | >2px | >3px | Avg/px |
|---|---|---|---|---|
| SGM | 16.41% | 9.69% | 6.66% | 1.34 |
| Gauss | 7.72% | 3.76% | 2.51% | 0.78 |
| Diffusion Based | 8.24% | 2.74% | 1.53% | 0.74 |
| **Riverbed** | **2.60%** | **1.06%** | **0.73%** | **0.45** |

To verify the effectiveness of our riverbed method on aerial stereo pairs, we used aerial imagery and LiDAR data for Guangzhou, China. The ground resolution of the aerial images was about 3.2cm, and the LiDAR point density was approximately 30 points/m$^2$. According to the selection principle of the window size shown in Subsection 4.1, since the LiDAR points percentage was close to 3%, we set the window size at 5×5.

Figure 8 shows a few example depth maps obtained with SGM and different cost volume optimization methods. The upper and lower depth maps are the matching results for the flat land and the urban area, respectively. The disparity map generated by SGM was generally accurate, except for some gross errors in difficult matching areas, such as shadows, trees, and buildings. It can be seen in Figure 8 that the gross errors of the depth map gradually decreased from left to right and that our riverbed method produced the best results.

As the direction of sunlight changed with time, the shadow of the lamp post also was different on the stereo image. This phenomenon easily could have caused the pixels on the shadow to be regarded as matching points in the dense matching process and further led to the appearance of striped elevation lines on the flat basketball court, as shown in the SGM, Gauss, and diffusion-based depth maps in Figure 8. The surface shadows of trees on the road also caused the shadow areas on the road to fail to match. However, the introduction of LiDAR constraints effectively reduced the false interference of shadows on dense matching, as shown in the results of our riverbed method in Figure 8 (e).

In addition to shadow areas, low-texture and repeated texture areas, such as trees, grass, and building areas caused mismatches as shown in the second, third, and fourth column of Figure 8; but our riverbed method produced a higher matching rate, as shown in the fifth column of Figure 8. In summary, our riverbed method definitely aided in improving the matching results in shadows, low texture, and repeated texture regions, which is an important challenge for current stereo matching algorithms.

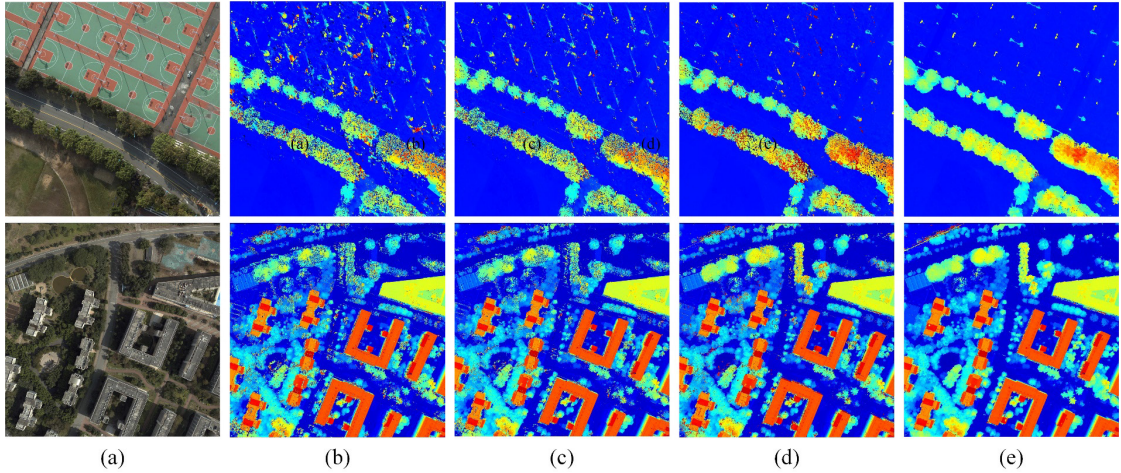

(a)      (b)      (c)      (d)      (e)

**Figure 8.** Experimental results on Guangzhou Stereo Dataset. (a) Left rectified image; (b) SGM results; (c) Gauss results; (d) Diffusion Based results; (e) Our riverbed results

The input and output point clouds of our riverbed method in Figure 9 are the same viewing angle, which is approximately vertical to the ground. As shown in Figure 9(a), the input/LiDAR point clouds are sparse and unevenly distributed. As shown in Figure 9(b), the output point clouds of our riverbed method are dense and accurate, such as street lights, basketball hoops, and building boundaries are clearly visible. More notably, the input/LiDAR point clouds are missing in local areas, while the output point clouds of our riverbed method recovered most of the missing point clouds. Therefore, as shown in Figure 8 and Figure 9, the complementary fusion of both LiDAR data and stereo images can produce the high-precision and high-density point clouds by our riverbed method.

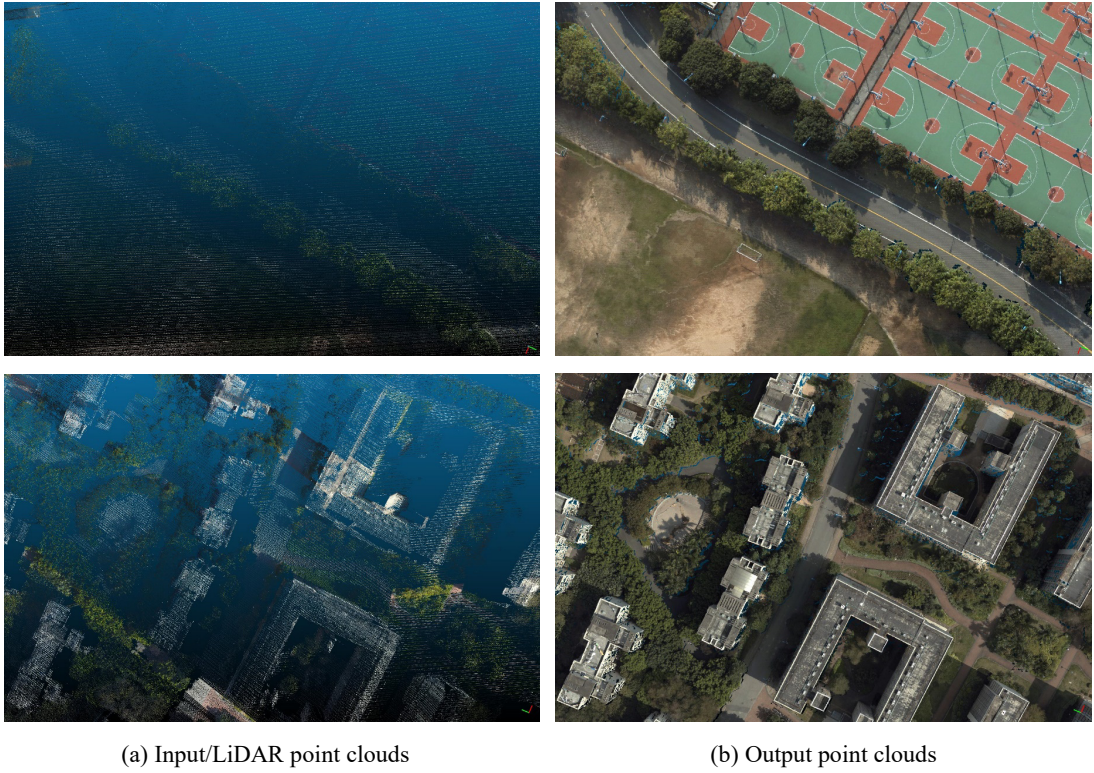

(a) Input/LiDAR point clouds          (b) Output point clouds

**Figure 9.** The input and output point clouds of our riverbed method.

## 4.4. Migration analysis

To prove that our riverbed method produced good results even with different stereo matching algorithms, we evaluated it using the AD-Census algorithm instead of the SGM algorithm used above. AD-Census consists of four steps: 1) AD-Census cost initialization, 2) cross-based cost aggregation, 3) scanline optimization, and 4) multi-step disparity refinement, which balances the matching effect and calculation efficiency well. AD-Census is integrated into the real-time stereo matching of Intel RealSense D400 due to extremely high matching efficiency (Keselman et al., 2017). Based on AD-Census (Li, 2021), we applied the LiDAR constraint cost volume before scanline optimization and did not adjust any parameters, which was consistent with the optimization strategy of the SGM algorithm.

As for any previous experiments, we randomly sampled the sparse inputs described in Subsections 4.2 and 4.3 and obtained an average LiDAR points percentage below 5%. Table 4 compares AD-Census and its cost volume optimization using sparse input cues on Middlebury 2014 (left) and KITTI 2015 (right), which represent the simulated datasets and the real datasets, respectively. For both datasets, our riverbed method was significantly better than standard AD-Census in all the metrics, and the cost volume optimization we propose produced the best results. In Middlebury 2014, our riverbed method produced only a few tenths of the amount of outliers and the average error of that of standard AD-Census. In particular, we reduced the amount of outliers> 3 from 32.70% to 0.25%, and the average error was reduced from 19.81 to 0.3. On KITTI 2015, our riverbed method also dramatically improved the amount of outliers and the average error. In particular, when compared to standard AD-Census, our riverbed method reduced the amount of outliers >1, >2, and >3 by more than 90%; and when compared to the two other methods, our riverbed method reduced the amount of outliers by more than 50%. More importantly, among all the methods, only our riverbed method reduced the average error to a sub-pixel level.

Our evaluations with AD-Census and the aforementioned SGM confirm that our riverbed method can be regarded as a general-purpose LiDAR-guided stereo matching that is capable of significant improvements using different stereo matching algorithms.

Table 4. The matching accuracy of standard and different LiDAR constraint AD-Census on the Middlebury and KITTI datasets.

| Method | Middlebury 2014 | | | | KITTI 2015 | | | |
|---|---|---|---|---|---|---|---|---|
| | >1px | >2px | >3px | Avg/px | >1px | >2px | >3px | Avg/px |
| AD-Census | 48.44% | 37.68% | 32.70% | 19.81 | 29.32% | 21.06% | 15.64% | 3.34 |
| Gauss | 5.43% | 4.43% | 4.26% | 3.56 | 6.85% | 4.35% | 3.63% | 1.24 |
| Diffusion Based | 5.53% | 3.19% | 3.03% | 2.77 | 7.05% | 3.56% | 2.89% | 1.18 |
| Riverbed | 0.94% | 0.29% | 0.25% | 0.30 | 2.88% | 1.03% | 0.75% | 0.53 |

## 5. Conclusions

In this paper, we proposed a novel method that utilizes a riverbed enhancement function for optimizing the cost volume of LiDAR projection points and their homogeneous pixels to extend the spatial consistency constraints of LiDAR projection points. Our experiments verified that our riverbed method is suitable for various stereo datasets, such as indoor, street, aerial, and satellite images. Furthermore, the qualitative and quantitative evaluations we conducted confirmed that our riverbed method was superior to two state-of-the-art methods, especially in reducing mismatches in difficult matching areas and refining the boundaries of objects or buildings.

Window size selection a fully automatic process in our method that is adaptive to the percentage of the LiDAR projection points on the image. When the LiDAR points percentage was 0.16%, the matching accuracy of our method achieved a sub-pixel level, while the original stereo matching algorithm was 3.4 pixels. To show the adaptability of our method, we applied the AD-Census algorithm in place of our original SGM algorithm and achieved great improvement effects, which further enhanced the application value and potential use of our riverbed method. For example, in low light conditions, as shown in the first column of Figure 7, our LiDAR-guided stereo matching strategy produced better results than the original stereo matching strategy, which will be useful for improving the 3D perception ability of autonomous vehicles.

Our riverbed method takes advantage of the high fidelity of LiDAR data, which requires that all LiDAR projection points are correct. Therefore, LiDAR data and images need to be strictly registered and the LiDAR point clouds must be as accurate as possible, which fortunately can be completed in a preprocessing step in most cases. As the amount of available data increases, we intend to integrate both the registration of LiDAR data and images and the elimination of outliers into our framework in future work.

## References

Arnold, E., Al-Jarrah, O.Y., Dianati, M., Fallah, S., Oxtoby, D., Mouzakitis, A., 2019. A Survey on 3D Object Detection Methods for Autonomous Driving Applications. IEEE Transactions on Intelligent Transportation Systems 20, 3782-3795.

Barron, J.T., Poole, B., 2016. The fast bilateral solver, European Conference on Computer Vision. Springer, pp. 617-632.

Cui, Y., Chen, R., Chu, W., Chen, L., Tian, D., Li, Y., Cao, D., 2021. Deep learning for image and point cloud fusion in autonomous driving: A review. IEEE Transactions on Intelligent Transportation Systems, 1-18.

Eldesokey, A., Felsberg, M., Khan, F.S., 2019. Confidence propagation through cnns for guided sparse depth regression. IEEE transactions on pattern analysis and machine intelligence 42,

2423-2436.

Gandhi, V., Cech, J., Horaud, R., Ieee, 2012. High-Resolution Depth Maps Based on TOF-Stereo Fusion, 2012 IEEE International Conference on Robotics and Automation, pp. 4742-4749.

Gao, X., Shen, S., Zhu, L., Shi, T., Wang, Z., Hu, Z., 2020. Complete Scene Reconstruction by Merging Images and Laser Scans. IEEE Transactions on Circuits and Systems for Video Technology 30, 3688-3701.

He, K., Sun, J., Tang, X., 2012. Guided image filtering. IEEE transactions on pattern analysis and machine intelligence 35, 1397-1409.

Hirschmuller, H., 2005. Accurate and efficient stereo processing by semi-global matching and mutual information, in: Schmid, C., Soatto, S., Tomasi, C. (Eds.), 2005 IEEE Computer Society Conference on Computer Vision and Pattern Recognition, Vol 2, Proceedings, pp. 807-814.

Hirschmuller, H., 2008. Stereo processing by semiglobal matching and mutual information. IEEE Transactions on pattern analysis and machine intelligence 30, 328-341.

Huang, X., Qin, R., Xiao, C., Lu, X., 2018. Super resolution of laser range data based on image-guided fusion and dense matching. ISPRS Journal of Photogrammetry and Remote Sensing 144, 105-118.

Huang, X., Wan, X., Peng, D., 2020. Robust Feature Matching with Spatial Smoothness Constraints. Remote Sensing 12, 3158.

Huber, D., Kanade, T., 2011. Integrating lidar into stereo for fast and improved disparity computation, 2011 International Conference on 3D Imaging, Modeling, Processing, Visualization and Transmission. IEEE, pp. 405-412.

Katz, S., Tal, A., Basri, R., 2007. Direct visibility of point sets. ACM Transactions on Graphics 26.

Keselman, L., Woodfill, J.I., Grunnet-Jepsen, A., Bhowmik, A., Ieee, 2017. Intel (R) RealSense (TM) Stereoscopic Depth Cameras, 2017 IEEE Conference on Computer Vision and Pattern Recognition Workshops, pp. 1267-1276.

Kopf, J., Cohen, M.F., Lischinski, D., Uyttendaele, M., 2007. Joint bilateral upsampling. ACM Transactions on Graphics 26, 96–es.

Landeschi, G., 2018. Rethinking GIS, three-dimensionality and space perception in archaeology. World Archaeology 51, 17-32.

Li, Y., 2020. SemiGlobalMatching, https://github.com/ethan-li-coding/SemiGlobalMatching.

Li, Y., 2021. AD-Census, https://github.com/ethan-li-coding/AD-Census.

Liu, K., Ma, H., Ma, H., Cai, Z., Zhang, L., 2020. Building Extraction from Airborne LiDAR Data Based on Min-Cut and Improved Post-Processing. Remote Sensing 12, 2849.

Maltezos, E., Kyrkou, A., Ioannidis, C., 2016. LIDAR vs dense image matching point clouds in complex urban scenes, Fourth International Conference on Remote Sensing and Geoinformation of the Environment (RSCy2016). International Society for Optics and Photonics, p. 96881P.

Mandlburger, G., Wenzel, K., Spitzer, A., Haala, N., Glira, P., Pfeifer, N., 2017. IMPROVED TOPOGRAPHIC MODELS VIA CONCURRENT AIRBORNE LIDAR AND DENSE IMAGE MATCHING, ISPRS Annals of Photogrammetry, Remote Sensing & Spatial Information Sciences, pp. 259-266.

Mei, X., Sun, X., Zhou, M., Jiao, S., Wang, H., Zhang, X., 2011. On building an accurate stereo matching system on graphics hardware, 2011 IEEE International Conference on Computer Vision Workshops (ICCV Workshops). IEEE, pp. 467-474.

Menze, M., Geiger, A., Ieee, 2015. Object Scene Flow for Autonomous Vehicles, 2015 IEEE Conference on Computer Vision and Pattern Recognition, pp. 3061-3070.

Menze, M., Heipke, C., Geiger, A., 2018. Object Scene Flow. ISPRS Journal of Photogrammetry and Remote Sensing 140, 60-76.

Min, D., Lu, J., Do, M.N., 2012. Depth video enhancement based on weighted mode filtering. IEEE Transactions on Image Processing 21, 1176-1190.

Nefti-Meziani, S., Manzoor, U., Davis, S., Pupala, S.K., 2015. 3D perception from binocular vision for a low cost humanoid robot NAO. Robotics and Autonomous Systems 68, 129-139.

Nickels, K.M., Castano, A., Cianci, C., 2003. Fusion of lidar and stereo range for mobile robots, in: Nunes, U., deAalmeida, A.T., Bejczy, A.K., Kosuge, K., Macgado, J.A.T. (Eds.), Proceedings of the 11th International Conference on Advanced Robotics 2003, Vol 1-3, pp. 65-70.

Patil, S., Comandur, B., Prakash, T., Kak, A.C., 2019. A new stereo benchmarking dataset for satellite images. arXiv:1907.04404.

Poggi, M., Pallotti, D., Tosi, F., Mattoccia, S., 2019. Guided stereo matching, Proceedings of the IEEE/CVF Conference on Computer Vision and Pattern Recognition, pp. 979-988.

Qin, R., 2017. Automated 3D recovery from very high resolution multi-view images Overview of 3D recovery from multi-view satellite images, ASPRS Conference (IGTF) 2017, pp. 12-16.

Remondino, F., Spera, M.G., Nocerino, E., Menna, F., Nex, F., 2014. State of the art in high density image matching. The photogrammetric record 29, 144-166.

Salvi, J., Pagès, J., Batlle, J., 2004. Pattern codification strategies in structured light systems. Pattern Recognition 37, 827-849.

Scharstein, D., Hirschmüller, H., Kitajima, Y., Krathwohl, G., Nešić, N., Wang, X., Westling, P., 2014. High-Resolution Stereo Datasets with Subpixel-Accurate Ground Truth, in: Jiang, X., Hornegger, J., Koch, R. (Eds.), Pattern Recognition, pp. 31-42.

Scharstein, D., Szeliski, R., 2002. A taxonomy and evaluation of dense two-frame stereo correspondence algorithms. International Journal of Computer Vision 47, 7-42.

Schenk, T., Csathó, B., 2002. Fusion of LIDAR data and aerial imagery for a more complete surface description. International Archives of Photogrammetry Remote Sensing and Spatial Information Sciences 34, 310-317.

Shivakumar, S.S., Mohta, K., Pfrommer, B., Kumar, V., Taylor, C.J., 2019. Real time dense depth estimation by fusing stereo with sparse depth measurements, 2019 International Conference on Robotics and Automation (ICRA). IEEE, pp. 6482-6488.

Tomasi, C., Manduchi, R., Ieee, 1998. Bilateral filtering for gray and color images, Sixth International Conference on Computer Vision, pp. 839-846.

Veitch-Michaelis, J., Muller, J.P., Storey, J., Walton, D., Foster, M., 2015. DATA FUSION OF LIDAR INTO A REGION GROWING STEREO ALGORITHM, in: Fuse, T., Nakagawa, M. (Eds.), Indoor-Outdoor Seamless Modelling, Mapping and Navigation, pp. 107-112.

Wang, R., Ferrie, F.P., 2015. Upsampling method for sparse light detection and ranging using coregistered panoramic images. Journal of Applied Remote Sensing 9, 095075.

White, J.C., Wulder, M.A., Vastaranta, M., Coops, N.C., Pitt, D., Woods, M., 2013. The Utility of Image-Based Point Clouds for Forest Inventory: A Comparison with Airborne Laser Scanning. Forests 4, 518-536.

Xiao, W., Mills, J., Guidi, G., Rodríguez-Gonzálvez, P., Gonizzi Barsanti, S., González-Aguilera,

D., 2018. Geoinformatics for the conservation and promotion of cultural heritage in support of the UN Sustainable Development Goals. ISPRS Journal of Photogrammetry and Remote Sensing 142, 389-406.

Yoon, K.J., Kweon, I.S., 2006. Adaptive support-weight approach for correspondence search. IEEE transactions on pattern analysis and machine intelligence 28, 650-656.

Zabih, R., Woodfill, J., 1994. Non-parametric local transforms for computing visual correspondence, European conference on computer vision. Springer, pp. 151-158.

Zhang, Y., Xiong, X., Zheng, M., Huang, X., 2015. LiDAR Strip Adjustment Using Multifeatures Matched With Aerial Images. IEEE Transactions on Geoscience and Remote Sensing 53, 976-987.

Zhang, Y., Zheng, Z., Luo, Y., Zhang, Y., Wu, J., Peng, Z., 2019. A CNN-Based Subpixel Level DSM Generation Approach via Single Image Super-Resolution. Photogrammetric Engineering & Remote Sensing 85, 765-775.

Zhou, K., Lindenbergh, R., Gorte, B., Zlatanova, S., 2020. LiDAR-guided dense matching for detecting changes and updating of buildings in Airborne LiDAR data. ISPRS Journal of Photogrammetry and Remote Sensing 162, 200-213.

Zhou, Y., Song, Y., Lu, J., 2018. Stereo Image Dense Matching by Integrating Sift and Sgm Algorithm, International Archives of the Photogrammetry, Remote Sensing and Spatial Information Sciences, p. 3.

Zomet, A., Peleg, S., 2002. Multi-sensor super-resolution, Sixth IEEE Workshop on Applications of Computer Vision, 2002.(WACV 2002). Proceedings. IEEE, pp. 27-31.